\documentclass[review]{elsarticle}

\usepackage{lineno}

\usepackage{helvet} 
\usepackage{courier}  

\usepackage{cite} 

\usepackage{amsfonts}
\usepackage{latexsym}
\usepackage{mathrsfs}
\usepackage{url}
\usepackage{algorithm}
\usepackage[noend]{algpseudocode}
\usepackage{multirow}
\usepackage{xcolor}
\usepackage{natbib}
\usepackage{verbatim}

\usepackage{amsmath}
\usepackage{amssymb}
\usepackage{bm}

\usepackage[misc]{ifsym}

\usepackage{graphicx}  
\journal{Journal of \LaTeX\ Templates}

\biboptions{authoryear}

\begin{document}

\begin{frontmatter}

\title{Neural Data-to-Text Generation with Dynamic Content Planning}

\author{Kai Chen$^\dag$\corref{equalContribution}}
\author{Fayuan Li$^\ddag$\corref{equalContribution}}
\cortext[equalContribution]{These authors have contributed equally to this work.}

\author{Baotian Hu$^\dag$\Letter}
\ead{hubaotian@hit.edu.cn}
\author{Weihua Peng$^\dag$}
\author{Qingcai Chen$^\dag$}
\author{Hong Yu$^\S$}




\address{$^\dag$Harbin Institute of Technology (Shenzhen)}
\address{$^\ddag$Beijing Baidu Netcom Science Technology Co., Ltd.}
\address{$^\S$University of Massachusetts Lowell}





\begin{abstract}
Neural data-to-text generation models have achieved significant advancement in recent years. However, these models have two shortcomings: the generated texts tend to miss some vital information, and they often generate descriptions that are not consistent with the structured input data. To alleviate these problems, we propose a {\bf N}eural data-to-text generation model with {\bf D}ynamic content {\bf P}lanning, named {\bf NDP} for abbreviation. The NDP can utilize the previously generated text to dynamically select the appropriate entry from the given structured data. We further design a reconstruction mechanism with a novel objective function that can reconstruct the whole entry of the used data sequentially from the hidden states of the decoder, which aids the accuracy of the generated text. Empirical results show that the NDP achieves superior performance over the state-of-the-art on ROTOWIRE dataset, in terms of relation generation (RG), content selection (CS), content ordering (CO) and BLEU metrics. The human evaluation result shows that the texts generated by the proposed NDP are better than the corresponding ones generated by NCP in most of time. And using the proposed reconstruction mechanism, the fidelity of the generated text can be further improved significantly.
\end{abstract}

\begin{keyword}
Data-to-Text, Dynamic Content Planning, Reconstruction Mechanism
\end{keyword}

\end{frontmatter}

\section{Introduction}

Language generation has been applied in many NLP applications, such as machine translation~\citep{Bahdanau:2014:arXiv}, text summarization~\citep{see_get_2017} and dialog system~\citep{lifengshang}. Unlike the above fields which take the text as input, data-to-text generation aims to produce informative, fluent and coherent multi-sentences descriptive text from the given structured data such as a table of sport game statistics~\citep{Robin94revision}, weather forecast data~\citep{belz_2008}, and so on.

Generally, data-to-text generation needs to tackle two major problems~\citep{Kukich:1983,McKeown:1985}: {\em what to say}, i.e., what data should be covered in the output text and {\em how to say}, i.e., how to convey the information using grammatically and logically corrected text. Most of the traditional work addresses these two issues via using different isolated modules with domain expert knowledge. On one hand, constructing a data-to-text generation system in this way is time-consuming and laborious. On the other hand, these systems are difficult to be extended to other domains.

Due to the recent fundamental advancements on neural language generation and representation~\citep{Bengio:2003:JMLR}, neural network based approaches have drawn increasing attentions such as ~\citet{nie2018operations},~\citet{ncp}, and ~\citet{wiseman2017-challenges}. Our work also falls into this direction. Unlike the traditional approach, neural network based models can be constructed almost from scratch with an end-to-end fashion. These models are usually based on the encoder-decoder framework, which is mainly borrowed from neural sequence-to-sequence models~\citep{see_get_2017,Sutskever:2014:NIPS,Bahdanau:2014:arXiv}. Although some recent work~\citep{ncp,wiseman2017-challenges} demonstrates that deep models perform much better than traditional approaches on maintaining inter-sentential coherence and a more reasonable ordering of the selected facts in the output text, neural data-to-text generation perform much worse on avoiding redundancy and being faithful to the input without using explicit data selecting and ordering.

To augment the neural models, ~\citet{ncp} propose an explicit content selection and planning model, which can select the data and their order before text generation. Their model is divided into two stages. The explicit content selection and planning are independent of the text generation and the semantic information embedded in the text do not participate in the content selection and planning. Ideally, the data should be dynamically selected in the process of text generation, which can make full use of the semantic information of the generation history. Conversely, the appropriate selected data can benefit to the text generation. In a word, the dynamic content selection and planning can benefit both "{\em what to say}" and "{\em how to say}". However, to the best of our knowledge, there is no work incorporating dynamic content selection and planning while generating text. Our work aims to energize the neural data-to-text generation model to dynamically select appropriate content from the given structured data with a novel dynamic planning mechanism.  

Some recent work focuses on improving the ability of content selection in the encoder part such as~\citet{ncp} and~\citet{nie2018operations} or the intermediate parts such as copy and coverage mechanism, the importance of the end part i.e., text decoder, lacks deep investigation. \citet{rec_tu} show that the reconstruction mechanism on the top of the decoder can improve the performance of machine translation significantly. Intuitively, a well-designed reconstruction mechanism can encourage the decoder to take more important information from the encoder side. Unlike ~\citet{wiseman2017-challenges} which reconstructs some specific fields (i.e., value and entity) of the data entry using a convolutional classifier, we use another recurrent neural network to reconstruct the whole data entry sequentially with a novel objective function.

The contribution of our work can be summarized as follows:
  \begin{itemize}
   \item We propose a novel differentiable dynamic content planning mechanism that can make full use of the previously generated history and the importance of the data self to decide which data should be used in the next step. The proposed dynamic content planning mechanism can be easily integrated with the encoder-decoder framework and it has its own objective function.
   \item To ensure the decoder generates text as accurate as possible, we further design a novel record reconstruction mechanism with a well-designed objective function, to encourage the decoder to take more accurate information from the encoder side. 
   
   \item Finally, we construct a novel {\em N}eural data-to-text generation model with proposed {\em D}ynamic content {\em P}lanning mechanism, named {\em NDP} for abbreviation. We experimentally evaluate the induced model on the challenging benchmark dataset ROTOWIRE ~\citep{wiseman2017-challenges}. The results show that NDP significantly improves the adequacy of generated text and achieves superior performance over state-of-the-art neural data-to-text systems in terms of relation generation (RG), content selection (CS), content ordering (CO) and BLEU metrics. The human evaluation result via using Best-Worst Scaling(BWS) technique~\citep{bws}, shows that the texts generated by the proposed NDP are much better than the corresponding ones generated by NCP in most of the time. Using the proposed reconstruction mechanism in the training period, the fidelity (in terms of precision) of the generated text can be further improved with a large margin.  
   
   
  \end{itemize}

\section{Background: Static Content Planning}
\label{conent_selection_static_planning}
In this section, we briefly introduce the explicit content planning mechanism proposed by~\citet{ncp}, which is the basis of our work. We call it as {\em static content planning} because once the data and its order are acquired, it will not change during the text generation. 

The data-to-text generation can be defined as generating a document $Y=(y_1,...,y_T)$ from a given structured data $D$. For different tasks, the form of $D$ may be different. In our scenario, we take NBA basketball game report generation challenge ROTOWIRE~\citep{wiseman2017-challenges} as our task. Concretely, the $D$ is a extensive statistical table consisting of a number of records $r$, i.e., $D=\{r_j\}^{|r|}_{j=1}$. Each record $r_j$ has four fields, i.e., type(e.g., $r_{j,1}$=POINTS), entity(e.g.,$r_{j,2}$= Kevin\_Love), value(e.g.,$r_{j,3}$=20), and whether a player is in a home-team(H) or visiting-team(V)(e.g.,$r_{j,4}$=H).

For each record $r_j$, ~\citet{ncp} first lookup embedding matrix $\mathbf{E}\in \mathbb{R}^{|V| \times d}$ for each record's features and concatenate them together. And then, a nonlinear layer is used to get $\overrightarrow{{r}_j} $ as Eq.~\ref{eq1}. 
\begin{equation}
\overrightarrow{r_j} = \mathbf{ReLU}(\mathbf{W}_r\mathbf{E}([r_{j,1}, r_{j,2}, r_{j,3}, r_{j,4}])+ \mathbf{b}_r)
\label{eq1}
\end{equation}
where
\begin{itemize}
  \item $\mathbf{W}_r \in \mathbb{R}^{d \times 4d}$, $\mathbf{b}_r \in \mathbb{R}^{d \times 1} $ are parameters, $d$ is the dimension of embedding;
  \item $\mathbf{ReLU}$ is the the rectifier activation function.
\end{itemize}

The $\overrightarrow{r_j}$ is used to attend to other records and the vector $r_j^{att}$ is obtained with following equations. 
  \[ \alpha_{j,k} \propto \mathbf{exp}(\overrightarrow{r_{j}}^T \mathbf{W}_a \overrightarrow{r_{k}}), \]
  \[ c_j = \sum_{k\neq j} \alpha_{j,k} \overrightarrow{r_{k}}, \]
  \[ r_j^{att} = \mathbf{W}_{g}[\overrightarrow{r_j}; c_j],\]
where $\mathbf{W}_a \in \mathbb{R}^{d \times d} $, $\mathbf{W}_g \in \mathbb{R}^{d \times 2d} $ are parameters and $\sum_{k\neq j} \alpha_{j,k} = 1 $.

The $r_j^{att}$ is further used to select the information from $\overrightarrow{r_j}$ with Eq.~\ref{eq2} and $r_j^{cs}$ is obtained.
\begin{equation}
r_j^{cs} = \mathbf{Sigmoid}(r_j^{att})\odot \overrightarrow{r_j}
\label{eq2}
\end{equation}
where $\odot$ denotes the element-wise product.

The static content plan is a sequence of pointer $[z_1,..,z_k]$, the $z_i$ refer to the record index in the given structured data input. The long short-term memory neural network(LSTM) is used to output the static record plan sequentially. The first hidden state is initialized with the average of $r^{cs}$ i.e., $h_0=\frac{1}{n}\sum_{j=1}^{n}r_j^{cs}$. On step $i$, the input of the LSTM is the representation of the previous selected record $r_{z_{i-1}}^{cs}$. Then the output hidden state $h_i$ is used to attend to all the other records and get the ${P}_i(z_i=j|h_i)$ as Eq.~\ref{eq3}. 
\begin{equation}
{P}_i(z_i=j|h_i,r_j) \propto  exp(h_i^T\mathbf{W}_c{r}_j^{cs}) 
\label{eq3}
\end{equation}
where $\mathbf{W}_c$ is the parameter and $\sum_{j=1}^{n}{P}_i(z_i=j|h_i)=1$.

For training, suppose the gold static content plan has been obtained denoted as $[\mathring{z}_{1},...,\mathring{z}_{k}]$ (The gold static content plan extraction will be explained in {\em Experiments} section). The static content planning module can be trained via minimizing the loss function as Eq.~\ref{eq4}.
\begin{equation}
\mathcal{L}_{sp} = -\sum_{i=1}^k\mathbf{log}{P}_i(z_i=\mathring{z}_{i}|h_i,r_j)
\label{eq4}
\end{equation}

For inference, Pointer Network~\citep{poniternetwork} is explored to output the index of the selected record. The $z_i$ for step $i$ is predicted by Eq.~\ref{eq5}.
\begin{equation}
z_i =\mathop{\arg\max}_{j} \\\ {P}_i(z_i=j|h_i,r_j)
\label{eq5}
\end{equation}

\section{Neural Data-to-Text Generation with Dynamic Content Planning}
\begin{figure}[!tb]
\centering
\includegraphics[width=1\textwidth]{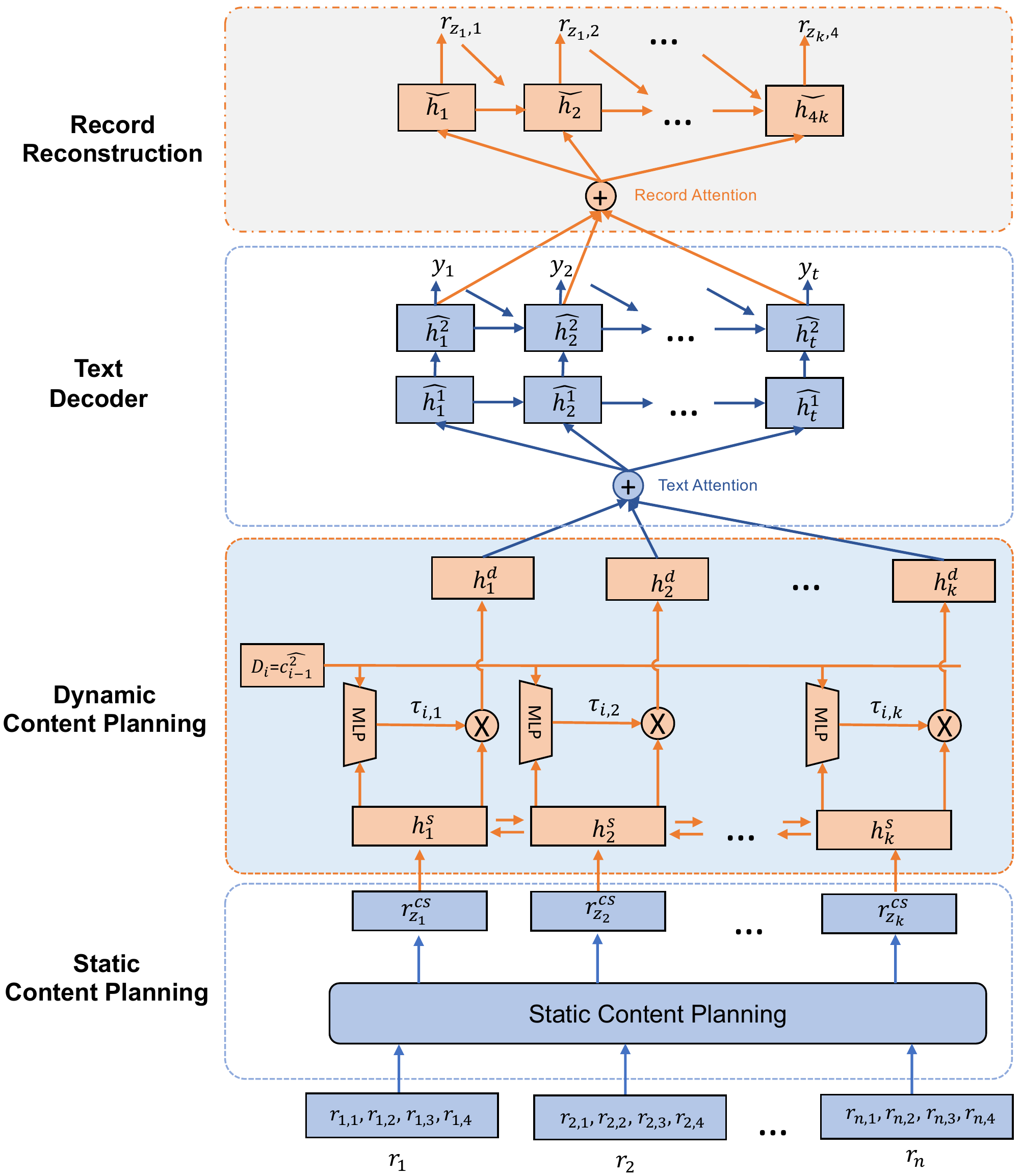}
\caption{The overall architecture of the proposed NDP. The reconstruction mechanism is optional, which is only used in training period.}
\label{our_model}
\end{figure}

The overall architecture of the proposed NDP with reconstruction mechanism is shown in Figure~\ref{our_model}. It consists of four components:
\begin{itemize}
  \item {\em Static Content Planning} that acquires the selected records and their order which will be fed into the following component. 
  \item {\em Dynamic Content Planning} is the novel part proposed in this paper. It will decide which record will play an important role in generating the next word according to the current state. 
  \item {\em Text Decoder} that generates word sequentially with attention~\citep{Bahdanau:2014:arXiv} and copy mechanism~\citep{gulcehre-etal-2016-pointing} from the dynamic content plan representation. 
  \item {\em Record Reconstruction} is the novel part proposed in this paper which encourages the decoder to generate more accurate information from the dynamic content plan representation. It will only be used in the training period. 
\end{itemize}


\subsection{\em Dynamic  Content Planning}
\label{dynamic_planning}
From the description of {\em Static Content Planning} section, we can see that it can select some records from the given structured data input, according to the importance of the record. However, it neglects the semantic information in the text when arranging their order(i.e., {\em Planning}). In this part, we propose a dynamic content planning mechanism that can make full use of previous generation history to decide which record will play an important role in generating the next word. 

Once the static record plan $[z_1,...,z_k]$ is acquired, it's corresponding representations $[r_{z_1}^{cs},...,r_{z_k}^{cs}]$ are fed into a bi-directional LSTM sequentially and the bi-directional context representations ${h}^{s}=[{{h}_1^{s}},...,{{h}_k^{s}}]$ are obtained as Eq.~\ref{eq6}. 

\begin{equation}
{h}^{s} = \overleftrightarrow{\mathbf{LSTM}}_{enc}(r_{z_1}^{cs},...,r_{z_k}^{cs})
\label{eq6}
\end{equation}
where $\overleftrightarrow{\mathbf{LSTM}}_{enc}(\cdot)$ denotes the bi-directional LSTM.

For using the information of previously generated words, the memory cell state of the text decoder is used to guide which record should be selected on the step $i$. Let $D_i=\widehat{c_{i-1}^2}$, the $\widehat{c_{i-1}^2}$ is the memory cell state of the second layer of the text decoder(we use the two-layer LSTM for text decoder, the detail described in {\em Text Decoder} section). And then, $D_i$ is concatenated with each state in ${h}^{s}$. By using a no-linear layer with a sigmoid activation function, the probability of selecting the record $j$ on step $i$ is calculated as Eq.~\ref{eq7}.
\begin{equation}
P_i(z_i^d=j|D_i,h_j^s) = \mathbf{Sigmoid}(\mathbf{W}_d[D_i;{{h}_j^{s}}])
\label{eq7}
\end{equation}
where $z_i^d$ is the dynamic pointer to which record should be used on step $i$. $\mathbf{W}_d \in \mathbb{R}^{1\times 2d}$ is the parameter and $[;]$ is the concatenation operation. 

And then, the $P_i(z_i^d=j|D_i,h_j^s)$ is normalized to ${\mathbf{\tau}_{i,j}}$ with respect to other records, as Eq.~\ref{eq8}.

\begin{equation}
{\mathbf{\tau}_{i,j}} =\frac{P_i(z_i^d=j|D_i,h_j^s)}{\sum_{l=1}^{k}{P_i(z_i^d=l|D_i,h_l^s)}}
\label{eq8}
\end{equation}
Finally, the dynamic content planning representation ${h}_i^{d}=[{{h}_{i,1}^{d}},...,{{h}_{i,k}^{d}}]$ on step $i$ is obtained by using Eq.~\ref{eq9}. 
\begin{equation}
{h}_{i,j}^{d} = {{\tau}_{i,j}}{{h}_j^{s}}
\label{eq9}
\end{equation}
Comparing with the static content planning representation ${h}^{s}$, it is obvious that the dynamic content planning representation will change dynamically for each decoding step.  


Formally, on each step, one record is selected. Suppose the record $j$ should be selected on step $i$(with regard to how to acquire which record should be selected for each step during training, it will be explained in {\em Experiments} section). To ensure the dynamic content planning mechanism selects the appropriate entry from $h^s$, we use the objective function as Eq.~\ref{eq10}.


\begin{equation}
\begin{split}
\mathcal{L}_{dp}^{i}=&logP_i(z_i^d=j|D_i,h_j^s)+ \\
 &\sum_{l\neq j}^k log(1-P_i(z_i^d=l|D_i,h_l^s))
\end{split}
\label{eq10}
\end{equation}
Hence, the loss of the dynamic content planning for generating the text $Y$ is the accumulative loss for each step, which can be formulated as Eq.~\ref{eq11}.
\begin{equation}
\begin{split}
\mathcal{L}_{dp}=-\frac{1}{T}\sum_{l=1}^T \mathcal{L}_{dp}^{l}
\end{split}
\label{eq11}
\end{equation}
It should be noticed that the dynamic content planning is trained with the loss $\mathcal{L}_{dp}$ in a supervised way. They should be regarded as a whole part. 


\subsection{\em Text Decoder}
\label{text_decoder}

The decoder is a two-layer LSTM denoted as ${\mathbf{LSTM}}_{dec}$. The initial states are intialized with the last state of the $\overleftrightarrow{\mathbf{LSTM}}_{enc}$. The generation starts from the given symbol $<$/begin$>$ and terminates when the symbol $<$/end$>$ is emitted or the maximum length is reached. On step $i$, the ${\mathbf{LSTM}}_{dec}$ takes the previous word $y_{i-1}$ and hidden state $\widehat{h_{i-1}^{2}}$ as input while outputting the hidden states $\widehat{{h_i}^{2}}$ and memory cell state $\widehat{{c_i}^{2}}$ as Eq~\ref{eq6-extra}.

\begin{equation}
(\widehat{{h_i}^{2}},\widehat{{c_i}^{2}}) = {\mathbf{LSTM}}_{dec}(\mathbf{E}(y_{i-1}) ,\widehat{h_{i-1}^{2}})
\label{eq6-extra}
\end{equation}

The ${{c_i}^{2}}$ is used for dynamic content planning and ${{h_i}^{2}}$ is used for attention mechanism~\citep{Bahdanau:2014:arXiv} as the following equations: 

\[ \beta_{ij} \propto exp(\widehat{h_i^{2}}^T \mathbf{W}_a h_{i,j}^{d}), \]
\[ a_{i} =\sum_{j=1}^{k} \beta_{ij} h_{i,j}^{d}, \]
\[ h_{i}^{att} = \mathbf{Tanh}(\mathbf{W}_d[\widehat{h_i^2};a_i]),\]
where $\mathbf{W}_a$ and $\mathbf{W}_d$ are parameters. 

For generating word, we exploited conditional copy mechanism~\citep{gulcehre-etal-2016-pointing}. The probability of generating words $y_l$ from vocabulary is computed as Eq~\ref{eq12}:
  \begin{equation}
  p_{gen}^{i}(y_l) = \mathbf{Softmax}_{y_l}(\mathbf{W}_oh_{i}^{att} + \mathbf{b}_y)
  \label{eq12}
  \end{equation}

The gate of whether copying or generating a word is computed as Eq.~\ref{eq13}:
 \begin{equation}
  p_{ga}^{i} = \mathbf{Sigmoid}(\mathbf{W}_{copy}h_{i}^{att} +\mathbf{b}_{copy})
  \label{eq13}
  \end{equation}
The final probability of emitting word $y_l$ at step $i$ is computed as Eq.~\ref{eq14}:
\begin{equation}
  p^i(y_l) = p_{ga}^i\sum_{y_l\leftarrow r_{z_k}}\beta_{i,k} + (1-p_{ga}^i)p_{gen}^{i}(y_l)
  \label{eq14}
\end{equation}

For training, suppose the reference text is $\mathring{Y}=(\mathring{y_1},...,\mathring{y_T})$. The loss of the text decoder can be calculated as Eq.~\ref{eq15}:
\begin{equation}
 \mathcal{L}_{lm} = -\frac{1}{T}\sum_{i=1}^T(\mathbf{log} p^i(\mathring{y_i})+\gamma_{l}\mathbf{log}(| p^i(\mathring{y_i})-\overline{p}_{1...T}|))
\label{eq15}
\end{equation}
where the $\overline{p}_{1...T}$ is the average probability of each word in the $\mathring{Y}$, i.e., $\overline{p}_{1...T}=\frac{1}{T}\sum_{i=1}^Tp^i(\mathring{y_i})$. The item $\gamma_{l}\mathbf{log}(| p^i(\mathring{y_i})-\overline{p}_{1...T}|)$ can be viewed as the a regularization which is designed for alleviating repetition problem. The hyper-parameter $\gamma_{l}$ can be chosen empirically.
  
\subsection{\em Record Reconstruction}
\label{Record_Reconstruction}
Unlike~\citet{wiseman2017-challenges} which divides the decoder hidden state into two parts and uses the convolutional neural network~\citep{Collobert:2011:JMLR} to predict the entities and values of records, we use another LSTM  with attention to reconstruct all the fields of the selected data sequentially, as shown in Fig.~\ref{our_model}. The reconstruction loss can be computed as Eq.~\ref{eq16}.
\begin{equation}
 \mathcal{L}_{rec}^{'} = -\frac{1}{4k}\sum_{i=1}^k \sum_{j=1}^{4}\mathbf{log}p_{rec}(r_{z_i,j})
\label{eq16}
\end{equation}
where $p_{rec}(r_{z_i,j})$ is the generating probability of the $j$th element in the record $r_{z_i}$. The $p_{rec}(r_{z_i,j})$ is computed with softmax function and the input of the proposed record reconstruction mechanism is the hidden states of text decoder, the reconstruction is performed by sequentially outputting the filed of the golden records.  

Most of the previous work uses objective function~\ref{eq16}. In our work, to ensure the decoder generates the information as accurate as possible, we design another loss item that encourages the decoder hidden states to incorporate information that can be used to reconstruct all the fields of the selected record. The extra loss item is formulated as Eq.~\ref{eq17}.
\begin{equation}
 \mathcal{L}_{reg} = -\frac{1}{4k}\sum_{i=1}^k (\sum_{j=1}^{4}(\mathbf{log}|(p_{rec}(r_{z_i,j})-\overline{p}_{rec,z_i}|)
 \label{eq17}
\end{equation}
where $\overline{p}_{rec,z_i}$ is the average probability of all elements in record $r_{z_i}$ which can be computed as Eq.~\ref{eq18}. 
\begin{equation}
 \overline{p}_{rec,z_i} = \frac{1}{4}\sum_{j=1}^{4}p_{rec}(r_{z_i,j})
 \label{eq18}
\end{equation}

The $\mathcal{L}_{reg}$ can be viewed as a regularization item that prevents the record reconstruction of LSTM from over-fitting on some high-frequency elements while neglecting other fields. Finally, the overall loss of the record reconstruction can be summarized as Eq.~\ref{eq19}.
\begin{equation}
 \mathcal{L}_{rec} = \mathcal{L}_{rec}^{'}+ \gamma_{r} \mathcal{L}_{reg}
 \label{eq19}
\end{equation}
where the hyper-parameter $\gamma_{r}$ can be chosen empirically.

\subsection{\em Training}
\label{Training}
To summarize, each component of the proposed NDP has its objective function. The NDP can be trained in an end-to-end fashion, minimizing the overall loss as Eq.~\ref{eq20} while using the reconstruction mechanism. 

\begin{equation}
\begin{split}
\mathcal{J}(\theta)=&\lambda_1\mathcal{L}_{sp}+\lambda_2\mathcal{L}_{dp} \\
 &+\lambda_3\mathcal{L}_{lm}+\lambda_4\mathcal{L}_{rec}
\end{split}
\label{eq20}
\end{equation}
where $\lambda_1,\lambda_2,\lambda_3$ and $\lambda_4$ are the hyper-parameters which can be chosen empirically.

\section{Experiments}
\subsection{\em Data}
We evaluated the proposed NDP on the ROTOWIRE \citep{wiseman2017-challenges}, a large scale NBA basketball game summaries, paired with the corresponding box- and line-score tables. Comparing with other datasets, it is more challenging, much larger and the summaries are professionally written, relatively well structured and long (337 words on average). The number of record types is 39, the average number of records is 628, the vocabulary size is 11.3K and the token count is 1.6M. Followed previous work, we trained on 3,398 summaries, tested on 728, and used 727 for validation.

\begin{table}
\centering
\footnotesize
\begin{tabular}{|p{2mm}p{11mm}p{22mm}p{17mm}p{5mm}|}
\hline
ID & Value  & Entity & Value & H/V  \\
\hline 1 & LeBron & LeBron\_James & F\_NAME & V \\
 2 & James & LeBron\_James & S\_NAME & V \\
 3 & Kevin & Kevin\_Love  & F\_NAME & V   \\
 4 & Love  & Kevin\_Love  & S\_NAME & V   \\
 5 & 25 & LeBron\_James  & PTS & V  \\
 6 & 14 & LeBron\_James   & AST & V   \\
 7 & Kevin & Kevin\_Love  & F\_NAME & V  \\
 8 & Love & Kevin\_Love  & S\_NAME & V  \\
 9& 20 & Kevin\_Love & PTS & V  \\
 10 & 11 & Kevin\_Love & REB & V   \\
 11 & Kyrie & Kyrie\_Irving & F\_NAME & V   \\
 12 &  Irving & Kyrie\_Irving & S\_NAME & V   \\
 13 & 3 & Kyrie\_Irving & FGM & V   \\
 14 & 17 & Kyrie\_Irving & FGA & V   \\
 15 & 8 & Kyrie\_Irving & PTS & V    \\
 ... & ...  & ... & ...  & ...    \\
\hline
\end{tabular}
\caption{An example of gold static content plan extracted from the text in Table~\ref{gold-dynamic}.}
\label{gold-static}
\end{table}

\subsection{\em The Gold Static and Dynamic Content Plan Extraction}
\label{record_extraction}
\noindent \textbf{Static Content Plan Extraction}. In the training period, the gold static content plan is needed. We adopted the tool developed by~\citet{ncp} to extract the gold record plan. For the training set, the information extraction (IE) system developed by~\citet{ncp}, was first used to identify entity and value pairs in the text. And then the type of the entity-value pair was predicted. For the pair in the same sentence, and if there was a record in the record bank with matching entities and values, the pair was assigned the corresponding type. By processing the sentences in the text sequentially, a sequence of records was acquired. Player names were divided into first name and surname; team records are also prepossessed to indicate the name of the team’s city and the team itself. Table~\ref{gold-static} shows an example of the partial gold static content plan extracted via the above methods for its corresponding text in Table~\ref{gold-dynamic}.

\begin{table}[ht]
\centering
\footnotesize
\begin{tabular}{|p{118mm}|}
\hline
{The$^1$ dynamic$^1$ duo$^1$ of$^1$ \textcolor{red}{LeBron$^1$} \textcolor{red}{James$^2$} and$^3$ \textcolor{red}{Kevin$^3$} continued$^5$ their$^5$ outstanding$^5$ early$^5$ -$^5$ season$^5$ play$^5$ Saturday$^5$ .$^5$ James$^5$ posted$^5$	a$^5$ \textcolor{red}{25$^5$} -$^6$ point$^6$ ,$^6$ \textcolor{red}{14$^6$} -$^7$	assist$^7$ double$^7$ -$^7$	double$^7$ ,$^7$ while$^7$ \textcolor{red}{Kevin$^7$} \textcolor{red}{Love$^8$} also$^9$ accomplished$^9$ the$^9$ feat$^9$ with$^9$ \textcolor{red}{20$^9$} and$^{10}$ \textcolor{red}{11$^{10}$} boards$^{11}$ .$^{11}$ The$^{11}$ stellar$^{11}$ production$^{11}$ helped$^{11}$ overcome$^{11}$ a$^{11}$ down$^{11}$ night$^{11}$ for$^{11}$ \textcolor{red}{Kyrie$^{11}$} \textcolor{red}{Irving$^{12}$} ,$^{13}$ who$^{13}$ drained$^{13}$ just$^{13}$ \textcolor{red}{3$^{13}$} of$^{14}$ his$^{14}$ \textcolor{red}{17$^{14}$} shot$^{15}$ attempts$^{15}$ ,$^{15}$ producing$^{15}$ a$^{15}$ season$^{15}$ -$^{15}$ low$^{15}$ \textcolor{red}{8$^{15}$} -$^{16}$ point$^{16}$ tally$^{16}$.....} \\
\hline
\end{tabular}
\caption{An example of text with gold dynamic Content Plan. The token with red color is the one with the matching value in Table~\ref{gold-static}. The superscript number of each token is the corresponding record ID.}
\label{gold-dynamic}
\end{table}

\noindent \textbf{Dynamic Content Plan Extraction}. 
To train the proposed NDP, we need the gold dynamic content plan to calculate the loss in Eq.~\ref{eq11}. We first matched the token of the text with the value of the record in the gold static content plan (Table~\ref{gold-static}) sequentially. If there was a matching value and the entity of the record was presented in the same sentence, the corresponding record would be assigned to the current token. Otherwise, the next token's matching record would be assigned to the current token. As shown in Table ~\ref{gold-dynamic}, to match the entity, we did some processing like splitting the entity into tokens and matching each one of them.

\subsection{\em Evaluation Metrics}
\label{evalu_metrics}

We used the extractive evaluation tools developed by ~\citet{ncp} to compare our model with competitor models. It employs an accurate IE system on the gold and automatic summaries. Let $\mathring{Y}$ be the gold text, and $Y$ the generated text. It consists of three metrics: {\em Relation generation (RG)} computes the precision and number of unique relations extracted from $Y$ that also appear in the given structured input data. {\em Content selection (CS)} computes the precision and recall of unique relations extracted from $Y$ matching those found in $\mathring{Y}$. {\em Content ordering (CO)} computes the normalized Damerau-Levenshtein Distance~\citep{Brill:2000} between the sequences of records extracted from $Y$ and that extracted from $\mathring{Y}$. Besides, we report BLEU scores and human evaluation results.

\subsection{\em Experiment Setup}
We compared the proposed models with the following competitor models. {\em TEMPL}: Template-based generator constructed by~\citet{wiseman2017-challenges} which creates a document consisting of eight template sentences: an introductory sentence (who won/lost), six player-specific sentences (based on the six highest-scoring players in the game), and a conclusion sentence. {\em Wise:} The best reported system in~\citep{wiseman2017-challenges}. {\em ED+JC:} vanilla encoder-decoder model with attention and joint copy mechanism~\citep{gu:2016}. {\em ED+CC:} vanilla encoder-decoder model with attention and conditional copy mechanism. {\em NCP+JC:} proposed by~\citet{ncp} with static content planning and joint copy mechanism. {\em NCP+CC:} proposed by~\citet{ncp} with static content planning and conditional copy mechanism. {\em OpAtt:} is operation guided attention-based network proposed by~\citet{nie2018operations}.

To ensure the fairness of comparison with competitor models especially the strong baselines i.e.,NCP+CC and NCP+JC, we used the same embedding and LSTM dimension(600). The setting of the static content planning and text decoder LSTM was the same as NCP. To boost the training, we used the pre-trained parameters of NCP to initialize the corresponding parts of NDP. For the NDP+rec, we continued to train the pre-trained NDP model by adding the record reconstruction mechanism on the top of the text decoder. Input feeding~\citep{luong:2015} was used for the text decoder. Standard fully batched RNN~\citep{Bahdanau:2014:arXiv} was used for record reconstruction. We applied dropout with a rate of 0.3. With the initial learning rate 0.15, learning rate decay 0.97, Adagrad optimizer was used. The hyper-parameters $\lambda_1,\lambda_2,\lambda_3,\lambda_4,\gamma_l$ and $\gamma_r$ were set to 1, 0.05, 1, 1, 0.05 and 0.05 respectively. BPTT~\citep{Mikolov:2010:INTERSPEECH} was used in text decoder and truncation size was set to 100. We set the batch size to 5 for training and the beam size to 5 for inference. The NDP model is implemented based on PaddlePaddle\footnote{http://www.paddlepaddle.org/}.

\subsection{\em Result}
\label{Result}

\begin{table*}[tp]
\centering
\scriptsize
\begin{tabular}{|p{10.2mm}|p{3.8mm}p{4mm}p{4mm}p{4mm}p{4mm}|p{5.4mm}|p{4mm}p{4mm}p{4mm}p{4mm}p{4mm}|p{5.4mm}| }
\hline
&\multicolumn{6}{c|}{\multirow{1}*{{ \textbf{Validation Set}}}} & \multicolumn{6}{c|}{\multirow{1}*{{ \textbf{Test Set}}}}\\
\cline{2-7} \cline{8-13} \multicolumn{1}{|c|}{\multirow{2}*{{Model}}} &\multicolumn{2}{c}{RG} &\multicolumn{2}{c}{CS} &{CO}& \multicolumn{1}{c|}{\multirow{2}*{{BU}}} &  \multicolumn{2}{c}{RG} &\multicolumn{2}{c}{CS} &{CO}& \multicolumn{1}{c|}{\multirow{2}*{{BU}}}\\
\cline{2-6} \cline{8-12} &\# & P(\%) &P(\%) & R(\%) & \multicolumn{1}{c|}{DL(\%)} & &\# & P(\%) &P(\%) & R(\%) & \multicolumn{1}{c|}{DL(\%)} &\\
\hline TEMPL & \textbf{54.29} & \textbf{99.92} & 26.61 & \textbf{59.16} & 14.42 & 8.51  & \textbf{54.23} & \textbf{99.94} & 26.99 & \textbf{58.16} & 14.92 & 8.46 \\
\hline Wise & 23.95 & 75.10 & 28.11 & 35.86 & 15.33 & 14.57  & 23.72 & 74.80 & 29.49 & 36.18 & 15.42 & 14.19\\
\hline ED+JC & 22.98 & 76.07 & 27.70 & 33.29 & 14.36 & 13.22 & -& -& -& -& -& -\\
\hline ED+CC & 21.94 & 75.08 & 27.96 & 32.71 & 15.03 & 13.31 & -& -& -& -& -& -\\
\hline  OpAtt& - & - & - & - & - & 14.96 & - & - & - & - & - & 14.74\\
\hline NCP+JC & 33.37 & 87.40 & 32.20 & 48.56 & 17.98 & 14.92 & 34.09 & 87.19 & 32.02 & 47.29 & 17.15 & 14.89\\
\hline NCP+CC & 33.88 & 87.51 & 33.52 & 51.21 & 18.57 & 16.19 & 34.28 & 87.47 & 34.18 & 51.22 & 18.58 & 16.50\\
\hline NDP & 33.83 & 89.22 & {35.44} & {53.50} & 19.91 & \textbf{16.81} & 34.65 & 89.3 & 35.46 & 52.98 & 19.47 & \textbf{16.70}\\
\hline NDP+rec & 31.63 & 89.87 & \textbf{37.26} & 52.16 & \textbf{20.67} & 16.67 & 32.43 & 89.3 & \textbf{36.85} & 51.65 & \textbf{20.29} & 16.38 \\
\hline
\end{tabular}
\caption{Automatic evaluation on ROTOWIRE validation and test sets using relation generation (RG) count (\#) and precision (P\%), content selection (CS) precision (P\%) and recall (R\%), content ordering (CO) in normalized Damerau-Levenshtein distance (DL\%), and BLEU(BU)}
\label{table-all}
\end{table*}
The left part of Table~\ref{table-all} summaries the results on the validation set. NDP outperforms all the neural competitor models (Wise, ED+JCC, ED+CC, NCP+JC, NCP+CC, and OpAtt). Especially, NDP outperforms the NCP which only used static content planning mechanism, irrespective of the copy mechanism being employed. The difference between NDP with NCP+CC is that NDP uses both the proposed dynamic content planning and static content planning while NCP+CC only uses static content planning, which indicates that it is the dynamic content planning which brings performance improvement. As for the template-based system (TEMPL), Table~\ref{table-all} shows that it achieves much higher performances on RG and the recall of CS, comparing with machine learning models. It is not surprising as TEMPL depends on the domain expert knowledge, which can be viewed as an upper-bound on content selection and relation generation for machine learning based models. Neural models especially our proposed NDP irrespective of the reconstruction mechanism perform much better on the precision of content selection, content ordering, and BLEU metrics.  


The NDP outperforms NCP+CC significantly on the precision (P\%) of RG (89.22 vs 87.51), both of them get the comparable performance (33.83 vs 33.88) on the number (\#) of RG. Recall from the previous section that RG(\#) is the number of unique relations extracted from $Y$ that also appear in the given structured data. Our proposed dynamic content planning mechanism takes the output of static content planning as input, which decides how many records the proposed NDP can use from the given structured data. That is the reason that there no significant difference in the performance of the number (\#) of RG. The NDP+rec is the NDP model trained with the proposed record reconstruction mechanism. Comparing with NDP, NDP+rec improves the precision of the relation generation(RG) (89.87 VS 89.22), content selection(CS) (37.26 vs 35.44) and content ordering(CO) (20.67 vs 19.91), while it sacrifices some recall performances on RG (31.63 vs 33.83), CS (52.16 vs 53.50 ) and BLEU (16.67 vs 16.81). In other words, the reconstruction mechanism improves the fidelity of the generated text but sacrifices some adequacy. We believe that the proposed reconstruction mechanism is useful for some tasks whose fidelity is much more important than adequacy such as weather reports and stock market summary generation. 

The results on the test set shown in Table~\ref{table-all} follow a pattern similar to the validation set. NDP achieves higher performance in all metrics including relation generation, content selection, content ordering, and BLEU, comparing with NCP+CC and other neural competitor models. The proposed reconstruction mechanism can further improve the fidelity of the generated text.

We also did the human evaluation for NCP+CC, NDP and NDP+rec on 50 randomly selected samples. We asked three raters to evaluate 50 randomly selected samples for NCP+CC, NDP, and NDP+rec. For each sample, we arranged the generated summaries into three pairs i.e (NCP+CC, NDP), (NCP+CC, NDP+rec) and (NDP, NDP+rec). Each pair was shown to three raters, who were asked to choose which summary was best and which was worst according to four criteria: Fluency (is the summary fluent?), Conciseness (does the summary avoid redundant information and repetitions?), Fidelity(does the summary avoid erroneous information), and Overall(overall quality of the summary). We adopt the Best-Worst Scaling (BWS) technique ~\citep{bws} to get the final result.
BWS was shown to be less labor-intensive and providing more reliable results as compared to rating scales ~\citep{kiritchenko2017best}. 
The score ranges from -1.0 (absolutely worst) to +1.0 (absolutely best).

\begin{table}[ht]
\centering
\scriptsize
\begin{tabular}{|p{15mm}|p{11mm}|p{17mm}|p{12mm}|p{10mm}|}
\hline
Model & Fluency  & Conciseness & Fidelity & Overall  \\
\hline NCP+CC & -0.77 & -0.77 & -0.4 & -0.7 \\
 NDP & 0.23 & 0.37 & 0.03 & 0.27 \\
 NDP+rec & 0.53  & 0.4 & 0.37 & 0.4   \\
\hline
\end{tabular}
\caption{The human evaluation result on 50 randomly selected samples by using the Best-Worst Scaling (BWS). }
\label{table-human}
\end{table}

The human evaluation result is present in Table~\ref{table-human}. It shows that NDP achieves much higher scores on all four criteria than NCP+CC, which is in accordance with the results shown in Table~\ref{table-all} by using automatic metrics. It indicates that the generated summaries by NDP are better than the corresponding ones generated by NCP most of the time. The result in Table~\ref{table-human} also shows that the raters give higher scores to NDP+rec than NDP, although NDP+rec performs lower performance on the recall of relation generation and content selection. It indicates that the fidelity is much more import than the information adequacy of text for human raters. It is easily understood that the mistakes of the generated text are much easier to detect than the number of facts. Our further qualitative analysis indicates that the NDP tends to generate longer text (its length is closer to the human-written text) with fewer repetitions, comparing with the NCP. However, both of them make mistakes on the background knowledge which may be attributed to the lack of this knowledge in the input data. Due to the space limitation, a generated example is listed in Table~\ref{summary-gold-model} of {\bf Qualitative Example}. The text generated by NDP covers more records (blue color) than NCP+CC, which indicates that dynamic content planning can improve the adequacy of the generated text. NCP+CC performs not very well on avoiding the word repetitions (orange color). Both NDP and NCP+CC make some mistakes(red color). The text generated by NDP includes 6 mistakes, while NCP+CC includes 7 mistakes. Especially, the last two mistakes (two team's next games) of them are the background knowledge that is not included in the input data. These findings highlight the importance of introducing the background knowledge for generating rich descriptive text for sports game reports.

\subsection{\em Qualitative Example}

\begin{table}[ht]
\centering
\tiny
\begin{tabular}{|p{2mm}p{10mm}p{16mm}p{13mm}p{3mm}|p{2mm}p{8mm}p{15mm}p{8mm}p{3mm}|}
\hline
ID & Value  & Entity & Type & H/V & ID & Value  & Entity & Type & H/V  \\
\hline
1 & Golden\_State & Warriors & T-CITY & V &33 & 5 & Klay\_Thompson & FGM & V  \\
2 & Warriors & Warriors & T-NAME & V & 34 & Derrick & Derrick\_Favors & F\_NAME & H  \\
3 & 30 & Warriors & T-WINS & V   & 35 & Favors & Derrick\_Favors & S\_NAME & H\\
4 & 5 & Warriors & T-LOSSES & V  & 36 & 10 & Derrick\_Favors & FGM & H \\
5 & 116 & Warriors & T-PTS & V   & 37 & 16 & Derrick\_Favors & FGA & H\\
6 & Utah & Jazz & T-CITY & H   & 38 & 22 & Derrick\_Favors & PTS & H\\
7 & Jazz & Jazz & T-NAME & H   & 39 & 11 & Derrick\_Favors & REB & H\\
8 & 13 & Jazz & T-WINS & H   & 40 & Enes & Enes\_Kanter & F\_NAME & H\\
9 & 26 & Jazz & T-LOSSES & H   & 41 & Kanter & Enes\_Kanter & S\_NAME & H\\
10 & 105 & Jazz & T-PTS & H   & 42 & 13 & Enes\_Kanter & PTS & H\\
11 & 51 & Warriors & T-FG\_PCT & V   & 43 & 6 & Enes\_Kanter & FGM & H\\
12 & 52 & Warriors & T-FG3\_PCT & V  & 44 & 13 & Enes\_Kanter & FGA & H  \\
13 & 48 & Jazz & T-FG\_PCT & H   & 45 & 1 & Enes\_Kanter & FG3M & H\\
14 & 33 & Jazz & T-FG3\_PCT & H  & 46 & 1 & Enes\_Kanter & FG3A & H \\
15 & 44 & Jazz & T-REB & H   & 47 & 10 & Enes\_Kanter & REB & H \\
16 & 31 & Warriors & T-REB & V  & 48 & Rudy & Rudy\_Gobert & F\_NAME & H  \\
17 & Stephen & Stephen\_Curry & F\_NAME & V  & 49 & Gobert & Rudy\_Gobert & S\_NAME & H \\
18 & Curry & Stephen\_Curry & S\_NAME & V   & 50 & 16 & Rudy\_Gobert & PTS & H \\
19 & 10 & Stephen\_Curry & FGM & V  & 51 & 4 & Rudy\_Gobert & FGM & H \\
20 & 16 & Stephen\_Curry & FGA & V  & 52 & 9 & Rudy\_Gobert & FGA & H \\
21 & 4 & Stephen\_Curry & FG3M & V  & 53 & 8 & Rudy\_Gobert & FTM & H \\
22 & 9 & Stephen\_Curry & FG3A & V  &  54 & 11 & Rudy\_Gobert & FTA & H\\
23 & 27 & Stephen\_Curry & PTS & V  & 55 & 11 & Rudy\_Gobert & REB & H \\
24 & Draymond & Draymond\_Green & F\_NAME & V  & 56 & Gordon & Gordon\_Hayward & F\_NAME & H \\
25 & Green & Draymond\_Green & S\_NAME & V  &  57 & Hayward & Gordon\_Hayward & S\_NAME & H\\
26 & 15 & Draymond\_Green & PTS & V  & 58 & 17 & Gordon\_Hayward & PTS & H  \\
27 & 6 & Draymond\_Green & FGM & V  & 59 & 5 & Gordon\_Hayward & FGM & H \\
28 & 9 & Draymond\_Green & FGA & V  & 60 & 11 & Gordon\_Hayward & FGA & H \\
29 & 3 & Draymond\_Green & FG3M & V  &  61 & 1 & Gordon\_Hayward & FG3M & H\\
30 & 4 & Draymond\_Green & FG3A & V  & 62 & 4 & Gordon\_Hayward & FG3A & H \\
31 & Klay & Klay\_Thompson & F\_NAME & V  & 63 & 6 & Gordon\_Hayward & FTM & H \\
32 & Thompson & Klay\_Thompson & S\_NAME & V   & 64 & 6 & Gordon\_Hayward & FTA & H\\

\hline
\end{tabular}
\caption{An example of gold static content plan for its corresponding text in Table~\ref{summary-gold-model}.}
\label{gold-plan}
\end{table}

Table~\ref{gold-plan} shows the gold static content plan for the human-written text(Gold) in Table~\ref{summary-gold-model}. Table~\ref{summary-gold-model} lists the texts generated by NDP and NCP+CC models. We highlighted the text with blue color if it agrees with records in Table~\ref{gold-plan} and red if the text contradicts with records in Table~\ref{gold-plan}. We also use the orange color to highlight repetitions.

Table~\ref{summary-gold-model} shows that the text generated by NDP is much longer than NCP+CC. The text generated by NDP covers more records(blue color) than NCP+CC, which indicates that dynamic content planning can improve the adequacy of the generated text. NCP+CC performs not very well on avoiding the word repetitions(orange color), compared with the proposed NDP. Both NDP and NCP+CC make some mistakes(red color) on some facts. The text generated by NDP includes 6 mistakes, while NCP+CC includes 7 mistakes. Especially, the last two mistakes (two team's next games) of them are the background knowledge that is not included in the structured input data. These findings highlight the importance of introducing the background knowledge for generating rich descriptive text for sports game report, which will be our future work. 


\begin{table*}
\centering
\tiny
\begin{tabular}{|p{118mm}|}
\hline
{\textcolor{blue}{The Golden State Warriors ( 30 - 5 ) defeated the Utah Jazz ( 13 - 26 ) 116 - 105} on Wednesday at Energy Solutions Arena in Salt Lake City . The Warriors were the superior shooters in this game , going \textcolor{blue}{51 percent from the field and 52 percent from the three - point line , while the Jazz went 48 percent from the floor and just 33 percent from beyond the arc . While the Jazz out - rebounded the Warriors 44 - 31} , the Warriors made up for it by forcing the Jazz into 17 turnovers , while committing only 10 of their own . Stephen Curry was very tough to stop in this game , as \textcolor{blue}{he went 10 - for - 16 from the field and 4 - for - 9 from the three - point line to finish with a game - high of 27 points .} He also handed out 11 assists , notching his fourth double - double in his last five games . Over those last five outings , Curry is averaging 24 points and 11 assists per game , as he continues to have the hot hand . \textcolor{blue}{Draymond Green also had a strong showing in this one , scoring an efficient 15 points ( 6 - 9 FG , 3 - 4 3Pt ) .} Surprisingly , he only recorded one rebound though , which was a season - low for him . Klay Thompson only shot the ball eight times , \textcolor{blue}{converting on five} of those eight attempts . It was the least amount of shots he 's gotten off all year , but he was still able to add 12 points . The Warriors really spread around the minutes in this game with nine players recording more than 22 minutes . Despite the loss , the Jazz saw three players record double - doubles . \textcolor{blue}{Derrick Favors led the charge , going 10 - for - 16 from the field to score 22 points , while adding 11 rebounds as well .} He 's had 20 or more points in four out of his last five games , as he 's averaged 21 points per game over that span . \textcolor{blue}{Enes Kanter was another double - double contributor , with 13 points ( 6 - 13 FG , 1 - 1 3Pt ) and 10 rebounds of his own .} The Jazz received an impressive bench performance from \textcolor{blue}{Rudy Gobert , who recorded 16 points ( 4 - 9 FG , 8 - 11 FT ) and 11 rebounds . }It was arguably one of his best games of the season , as the 16 points were a season - high , and he also had a significant defensive presence with two steals and three blocks . \textcolor{blue}{Gordon Hayward finished with 17 points ( 5 - 11 FG , 1 - 4 3Pt , 6 - 6 FT ) as well .} The Warriors ' next game will be at home against the Miami Heat on Wednesday , while the Jazz will be at home against the Los Angeles Lakers on Friday .}  {\color{green}{\textbf{Gold}}}\\

\hline
\textcolor{blue}{The Golden State Warriors ( 30 - 5 ) defeated the host Utah Jazz ( 13 - 26 ) 116 - 105} on Wednesday at Energy Solutions Arena in Salt Lake City . The Warriors were the superior shooters in this game , going \textcolor{blue}{51 percent from the field and 52 percent from the three - point line , while the Jazz went 48 percent from the floor and just 33 percent from deep }. \textcolor{orange}{The Jazz also shot 48 percent from the field} and \textcolor{red}{52 percent from the three - point line },\textcolor{orange}{ while the Jazz finished at 48 percent from the floor and a meager 33 percent from deep .} The Jazz also forced \textcolor{red}{the Jazz} into \textcolor{red}{20 turnovers} , while committing only \textcolor{red}{12} of their own . The Warriors ' backcourt of Stephen Curry and Klay Thompson did the bulk of the work offensively . \textcolor{blue}{Curry finished with 27 points ( 10 - 16 FG , 4 - 9 3Pt , 3 - 3 FT )} and 11 assists , \textcolor{blue}{while Green posted 15 points ( 6 - 9 FG , 3 - 4 3Pt , 0 - 0 FT )} and three assists . He 's now pushed his season averages to 13 points and 6 rebounds on the year . Also coming off the bench , \textcolor{red}{David Green} chipped in with 10 points of his own . The Warriors ' next game will be on the road against \textcolor{red}{the Portland Trail Blazers on Friday} , while the Jazz will travel to \textcolor{red}{Minnesota to play the Timberwolves on Wednesday} . {\color{green}{\textbf{NCP+CC}}}\\
\hline
 {\textcolor{blue}{The Golden State Warriors ( 30 - 5 ) defeated the Utah Jazz ( 13 - 26 ) 116 - 105} on Wednesday at Energy Solutions Arena in Salt Lake City . The Warriors were the superior shooters in this game , \textcolor{blue}{going 51 percent from the field and 52 percent from the three - point line , while the Jazz finished at 48 percent from the floor and just 33 percent from deep .} The Jazz also forced \textcolor{red}{the Jazz} into \textcolor{red}{21 turnovers} , while committing only \textcolor{red}{nine} of their own . The Jazz were led by the duo of Stephen Curry and Klay Thompson . \textcolor{blue}{Curry finished with a game - high of 27 points ( 10 - 16 FG , 4 - 9 3Pt , 3 - 3 FT )} , while also adding 11 assists . It was his second double - double over his last three games , a stretch where he 's averaging 24 points and 12 assists . Draymond Green also had a strong showing , \textcolor{blue}{finishing with 15 points ( 6 - 9 FG , 3 - 4 3Pt , 0 - 0 FT )} and three assists . He 's now pushed his season averages to 12 points and 6 rebounds on the year . The only other Warrior to reach double figures in points was \textcolor{blue}{David Lee , who came off the bench for 10 points .} The Warriors ' next game will be on the road against the Golden State Warriors on Wednesday , while the Jazz will travel to Minnesota to play the Timberwolves on Wednesday . For the Jazz , it was a very tough loss for the Jazz . The Jazz were led by \textcolor{blue}{Derrick Favors , who posted a double - double of his own with 22 points ( 10 - 16 FG , 2 - 3 FT ) and 11 rebounds .} It was his second double - double in a row , as he 's combined for 54 points and 19 rebounds over his last two games . \textcolor{blue}{Rudy Gobert had a double - double of his own with 16 points ( 4 - 9 FG , 8 - 11 FT ) , 11 rebounds and three blocked shots .} It marked his third double - double in a row , a stretch where he 's averaging 17 points and 12 rebounds . \textcolor{blue}{Gordon Hayward chipped in with 17 points ( 5 - 11 FG , 1 - 4 3Pt , 6 - 6 FT ) of his own , while Enes Kanter chipped in with 13 points} ( \textcolor{red}{7 - 14 FG} , \textcolor{blue}{1 - 1 3Pt} ) and 10 rebounds as well . The Warriors ' next game will be on the road against \textcolor{red}{the New Orleans Pelicans on Friday} , while the Jazz will be at home against the \textcolor{red}{New Orleans Pelicans on Wednesday} .} {\color{green}{\textbf{NDP}}}\\

\hline
\end{tabular}
\caption{Example texts generated by NDP and NCP+CC models. The Gold denotes the human-written text.}
\label{summary-gold-model}
\end{table*}

\section{Related Work}



The key challenges in data-to-text generation are mainly contained two aspects: content or data selection and surface realization~\citep{Kukich:1983, McKeown:1985, Goldberg:1994}. The conventional approaches consider them as different individual parts. 

The content selection is addressed with the manual rules or obtained via deeply analyzing the alignment between the text and input data~\citep{Barzilay:2005,liang-etal-2009-learning,angeli-etal-2010-simple}. For surface realization, research has show that template-based approaches generally result in texts of high quality~\citep{Goldberg:1994, van-der-lee-etal-2017-pass}. However, it is time-consuming and difficult to make rules that satisfy all situations. Some researchers adopt the statistical machine translation based text generators such as ~\citet{wong:naacl07},~\citet{Belz:2009} and~\citet{Pereira2015TowardsAH}. However, these models are generally lower in performance~\citep{Reiter1995NLGVT}.

With the availability of large data-to-text datasets such as E2E NLG~\citep{novikova:2017} and ROTOWIRE data-set~\citep{wiseman2017-challenges}, there has been a growing interest in building neural network based systems. The main characteristic of these models is that there is no clear distinction between content selection and surface realization. 
Without the explicit content selection, the encoder-decoder frameworks~\citep{Sutskever:2014:NIPS, novikova:2017} perform much worse on metrics of content selection recall and factual output generation~\citep{wiseman2017-challenges, nie2018operations}. ~\citet{Sebastian:2018} apply the copy mechanism from neural summarization~\citep{see_get_2017} to improve the ability of content selection and~\citet{ncp} integrates the explicit content planning into neural models.~\citet{nie2018operations} proposes operation-guided attention to improve the fidelity of the generated text.

\section{Conclusion}
We propose a novel neural data-to-text generation model with dynamic content planning(NDP). To improve the fidelity of the generated text, we further propose a novel record reconstruction mechanism that encourages the decoder to use more accurate information from the encoder. Experimental results on the ROTOWIRE dataset show that the NDP achieves state-of-the-art performance over the strong baseline models in terms of relation generation, content selection, content order, and BLEU metrics. The human evaluation and qualitative analysis also demonstrate that the texts generated by the proposed NDP are much better than the corresponding ones generated by NCP in most of the time. However, the proposed NDP and previous models can not avoid making some mistakes on some facts generation, because there lacks necessary background knowledge in the input data. In the future, we will explore to use the extra knowledge graph to guide the text generation. In this way, we expect the generated text can include more background knowledge such as the team's previous performance or the information about future games and fewer mistakes.

\section{Acknowledgment}
This work is supported by Natural Science Foundation of China
(Grant No. 61872113, 61573118, U1813215, 61876052), Special
Foundation for Technology Research Program of Guangdong Province
(Grant No. 2015B010131010), Strategic Emerging Industry Development
Special Funds of Shenzhen (Grant No. JCYJ20170307150528934, JCYJ2017 0811153836555,
JCYJ20180306172232154), Innovation Fund of
Harbin Institute of Technology (Grant No. HIT. NSRIF.2017052).

\bibliography{main}

\end{document}